\documentclass[twocolumn]{article}

\usepackage{arxiv}

\RequirePackage{amsmath} 

\usepackage[utf8]{inputenc}
\usepackage[T1]{fontenc}

\usepackage{natbib}

\usepackage{hyperref} 
\usepackage{url}
\usepackage{amsfonts}
\usepackage{nicefrac}
\usepackage{microtype}     
\usepackage{cleveref}      
\usepackage{doi}
\usepackage{subcaption}
\usepackage{algorithm}
\usepackage[noend]{algpseudocode}
\usepackage{graphicx}
\usepackage{dsfont}

\usepackage{amsmath,amssymb,amsfonts}
\usepackage{dsfont}
\usepackage{subcaption}
\usepackage{graphicx}
\usepackage{url}
\usepackage{booktabs}

\usepackage{hyperref}
\usepackage{orcidlink}

\newtheorem{proposition}{Proposition}

\title{ConceptCF: Concept-based Counterfactuals for the Explainability of Time Series}
\date{}

\usepackage{authblk}

\newbox{\orcid}\sbox{\orcid}{\includegraphics[scale=0.06]{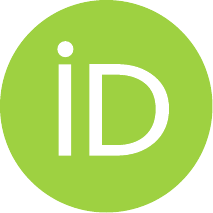}} 
\author[1,2]{%
	\href{https://orcid.org/0009-0004-9322-0674}{\usebox{\orcid}\hspace{1mm}Annemarie~Jutte\thanks{\texttt{a.m.p.jutte@saxion.nl}}}%
}
\author[1,2]{%
	\href{https://orcid.org/0000-0002-2760-6892}{\usebox{\orcid}\hspace{1mm}Faizan~Ahmed}%
}
\author[1]{%
	\href{https://orcid.org/0000-0002-2626-1837}{\usebox{\orcid}\hspace{1mm}Jeroen~Linssen}%
}
\author[2]{%
	\href{https://orcid.org/0000-0003-2436-1372}{\usebox{\orcid}\hspace{1mm}Maurice~van~Keulen}%
}
\affil[1]{Saxion University of Applied Sciences, Enschede, The Netherlands}
\affil[2]{University of Twente, Enschede, The Netherlands}

\renewcommand{\headeright}{}
\renewcommand{\undertitle}{}

\hypersetup{
pdftitle={ConceptCF},
pdfauthor={A.~Jutte et al.},
pdfkeywords={Explainable AI, Time series, Concept-based XAI, Counterfactuals},
}

\begin{document}
\twocolumn[{
  \begin{@twocolumnfalse}

	\maketitle

	\begin{abstract}
    This paper proposes ConceptCF, a method for counterfactual generation that operates on human-interpretable concepts. In high-stakes domains such as healthcare and predictive maintenance, artificial intelligence models can increase efficiency and safety. Explainability is key to ensure these models rely on causal relationships rather than spurious correlations. Counterfactual explanations identify minimal modifications that would change a model's predictions. Existing methods for time series operate on individual points or subsequences without ensuring interpretability of the mutations. ConceptCF instead modifies meaningful concepts. As a result we can provide explanations in terms of these concepts, for example ``the model's prediction would be `Sit' instead of `Walk' if you increase the scale of the movement''. In this paper, the concepts are constructed through time series decomposition, resulting in concepts such as scale, and frequency bands. Counterfactuals are generated using a genetic algorithm that optimizes the concept mutations. Evaluation against five state-of-the-art approaches demonstrates that ConceptCF consistently achieves top-tier performance across validity, confidence, proximity, sparsity and plausibility metrics.
	
    
    \end{abstract}

  \end{@twocolumnfalse}
  \vspace{2em}
}]




\section{Introduction}

Artificial Intelligence (AI) solutions are increasingly deployed in high-stakes domains such as healthcare and industry. However, many models operate under opaque conditions and may rely on spurious correlations rather than causal relationships~\cite{geirhos_shortcut_2020}. Explainable AI (XAI) addresses this concern with techniques to explain the behaviour of these models~\cite{adadi_peeking_2018}.

Counterfactual explanations~\cite{wachter_counterfactual_2017} have become popular in XAI because they explain not just why a prediction was made, but why another prediction was not made. This aligns with human reasoning, which is generally contrastive~\cite{miller_explanation_2019}. Unlike feature attribution methods, which demonstrate which feature values are important to a model's prediction, counterfactuals identify how features should be changed to alter this prediction.

In this paper, we focus on counterfactual explanations for time series data. The sequential nature of time series complicates counterfactual generation. Standard methods generally change features in isolation~\cite{wachter_counterfactual_2017}. However, in temporal data, modifying a single point can affect the semantic meaning of the other points in the sequence, complicating interpretation. Existing approaches, for example, preserve context by replacing entire sequences or subsequences~\cite{karlsson_locally_2020}. 

For counterfactuals to effectively explain model behaviour, they must clarify which data properties drive decisions. While such properties may be captured within subsequences, time series often display global characteristics, such as scale or periodicity, that cannot be captured in a few points. These characteristics are spread out across the signal. Additionally, modifying a subsequence does not guarantee interpretable changes. The changed properties may be unclear. Therefore, we instead modify concrete properties in the form of high-level human-interpretable patterns, so-called concepts.

We follow \citet{goyal_explaining_2020} to define concepts as features of a higher level than individual input features. Following this definition of concepts, we consider global patterns such as scale and periodicity~\cite{jutte_c-shap_2025} constructed using time series decomposition. Using these concepts, we propose Concept-based CounterFactuals (ConceptCF) for time series. We demonstrate ConceptCF using three decompositions: Fourier-based, Wavelet-based, and human-centered.

\begin{figure*}[ht]
    \centering
    \includegraphics[width=\linewidth]{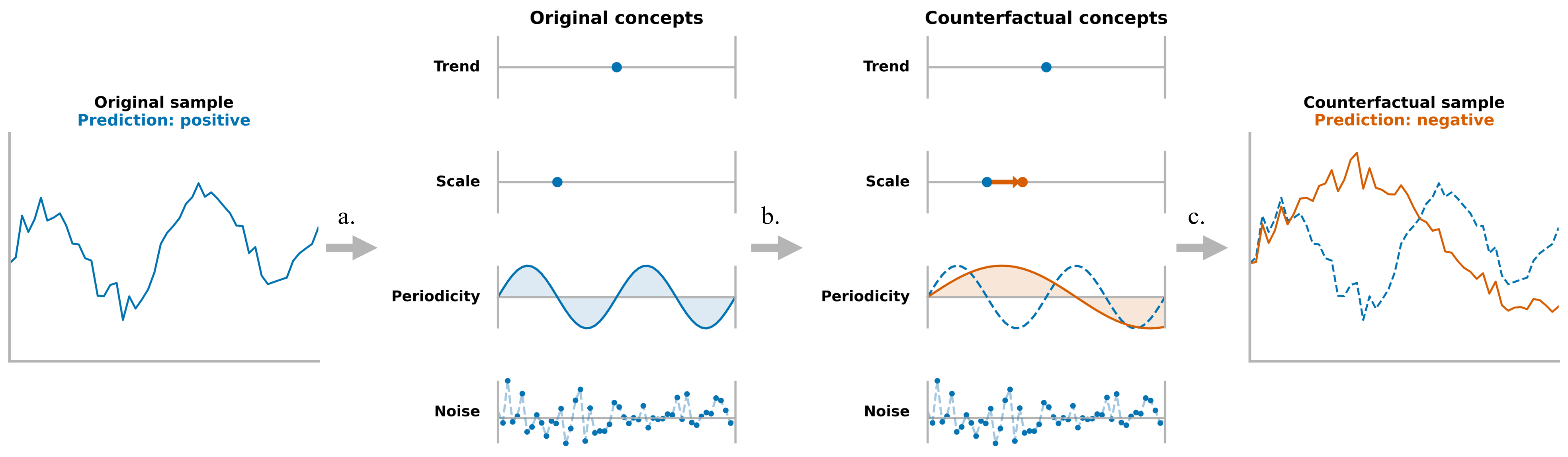}
    \caption{ConceptCF for time series. Unlike point-wise optimization, ConceptCF operates on underlying concepts to generate inherently interpretable counterfactuals. Concepts represent high-level features, expressed as scalars (e.g. scale) or sequences (e.g. periodicity). The figure illustrates an example where the prediction flips if the scale and wavelength are increased.}
    \label{fig:method}
\end{figure*}

ConceptCF operates in three stages, as illustrated in Fig.~\ref{fig:method}. First, concepts are extracted from a signal (see a. in Fig.~\ref{fig:method}). Second, counterfactuals are generated by perturbing these concepts rather than the raw signal (b.). Third, the perturbed concepts are mapped back to produce the counterfactual (c.). Because perturbations are applied at the concept level, explanations can be expressed in terms of concepts. For Figure~\ref{fig:method}, we can explain: “the prediction would change to negative if the scale and wavelength of the signal were increased”.

To summarize, we present the following contributions to the state of the art:
\begin{itemize}
    \item The formalization of the counterfactual problem for time series in the context of concept-based XAI.
    \item ConceptCF as a method for solving the concept-based counterfactual problem using a genetic algorithm.
    \item A demonstration of how concepts can be used for meaningful counterfactual generation.
    \item A validation demonstrating that ConceptCF performs competitively with existing counterfactual generation methods, in terms of validity, confidence, proximity, sparsity and plausibility.
    \item ConceptCF as a new tool to enhance our understanding of time series models, alongside these existing approaches.
\end{itemize}
The remainder of this paper is structured as follows. In Section~\ref{sec:related_work}, we discuss related work. In Section~\ref{sec:problem}, we define the optimization problem. In Section~\ref{sec:implementation}, we present our implementation. Finally, in Section~\ref{sec:experiments} we describe our validation experiments, and in Section~\ref{sec:results} we present their results.

\section{Related work}
\label{sec:related_work}

\subsection{Counterfactuals}
\label{sec:related_work_cfs}
Counterfactuals are generally defined as the solution to a multi-objective optimization problem. Objectives in existing methods include~\cite{guidotti_counterfactual_2024}: 
\begin{itemize}
    \item \textit{Validity}: The counterfactual should change the model prediction.
    \item \textit{Confidence}: The confidence of the model in the new counterfactual class should be high~\cite{cetina_counterfactual_2026}.
    \item \textit{Proximity}: The counterfactual should resemble the original sample.
    \item \textit{Sparsity}: For interpretable counterfactuals, only a small amount of features should be changed.
    \item \textit{Plausibility}: Meaningful counterfactuals should be within the distribution of the training data of the model.
\end{itemize}

Gradient-based methods~\cite{wachter_counterfactual_2017,wang_glacier_2024} efficiently solve the optimization problem requiring minimal model information, while model-agnostic approaches~\cite{dandl_multi-objective_2020,hollig_tsevo_2022} do not require any model information and allow non-differentiable objectives.

Applying counterfactual methods to time series results in two challenges due to their sequentiality. First, treating points independently risks generating out-of-distribution samples violating temporal dependencies~\cite{delaney_instance-based_2021}. Secondly, modifying individual points changes the semantic meaning of neighbouring points, complicating interpretation of the counterfactuals.

Some existing solutions replace full samples or subsequences with training data~\cite{karlsson_locally_2020}. Native Guide~\cite{delaney_instance-based_2021} uses a greedy approach, replacing subsequences with those from the nearest neighbour of an opposing class. Sub-SpaCE~\cite{refoyo_sub-space_2024} defines the subsequence substitution as an optimization problem, solved using a genetic algorithm. 

\citet{sulem_diverse_2022} do rely on a point-based optimization, but smooth perturbations across neighbouring points to mitigate out-of-distribution effects. Alternatively, LatentCF++~\cite{wang_learning_2021}, and its extension Glacier~\citep{wang_glacier_2024}, preserve the data distribution by perturbing the latent space of an autoencoder.

TSEvo~\cite{hollig_tsevo_2022} supports location- and frequency-based mutations, optimized by a genetic algorithm. Though not framed as such, frequencies can be seen as concepts, since they are high-level features. We mirror the frequency-based approach, redefine it within concept-based XAI and expand it with other concepts. Additionally, our exclusive focus on concepts simplifies interpretation.

\subsection{Concept-based XAI}

Concept-based XAI aims to increase the interpretability of explanations using `concepts' understandable for humans~\cite{poeta_concept-based_2023}. Rather than providing explanations in terms of input-level features which may not be directly interpretable to end users, such as individual time points in time, high-level, meaningful features are used.

Methods such as TCAV~\cite{kim_interpretability_2018}, CaCE~\cite{goyal_explaining_2020}, and ConceptSHAP~\cite{yeh_completeness-aware_2020} determine the attribution of concepts post-hoc. Other approaches are interpretable by design, such as prototypical neural networks~\cite{li_deep_2018} and Concept-Bottleneck Models (CBMs)~\cite{koh_concept_2020}.

However, concept attributions do not explain why a decision was made, this is a gap which counterfactuals address. To our knowledge, the only prior work applying concept-based XAI to counterfactuals for time series is TSEvo, and the CBM-based approach X-CHAR~\cite{jeyakumar_x-char_2023}. While X-CHAR considers concepts as meaningful subsequences, we consider a range of high-level features.

\section{Problem definition}
\label{sec:problem}

In this section, we define ConceptCF for time series. First, we define concepts for time series, next, we use the definition to pose our counterfactual problem. For conciseness, we limit our scope to univariate time series. However, the described method can be expanded to multivariate time series. 

\subsection{Concept definition}
Given a univariate time series $\mathbf{x} = (x_1, x_2, ..., x_n)\in\mathds{R}^n$, we define a concept $c_i$ as either a scalar in (complex) field $\mathds{C}$ or a sequence in field $\mathds{C}^j$, where $j\leq n$.

We define $S_\mathbf{x} = \{c_1, c_2, ..., c_m\}$ as a set of concepts constructed from time series $\mathbf{x}$. The concept set should be constructed from $\mathbf{x}$ using a function or algorithm $d(\cdot)$, such that $S_\mathbf{x} = d(\mathbf{x})$.

For  this paper, for the construction function $d(\cdot)$, we require the existence of a function $h(\cdot)$, such that $\mathbf{x} = h\left(S_\mathbf{x}\right) = h\left(d\left(\mathbf{x}\right)\right)$. In other words, the construction function should be invertible. This ensures we can reconstruct a meaningful signal from a set of perturbed concepts.

\subsection{Optimization problem}
ConceptCF poses counterfactual generation as an optimization problem. The concept set $S_{\mathbf{x}_\text{CF}}$ is optimized, which is a perturbed version of $S_{\mathbf{x}}$, which is constructed from $\mathbf{x}$. Once $S_{\mathbf{x}_\text{CF}}$ is optimized, $\mathbf{x}_{\text{CF}}$ can be constructed using $h(S_{\mathbf{x}_\text{CF}})$.

For the optimization, given $n_\text{c}$ target classes, predictive model $f(\cdot): \mathds{R}^n \to \mathds{R}^{n_\text{c}}$, and the predicted class $y_\mathbf{x} = \text{argmax}(f(\mathbf{x}))$, we formulate the problem in terms of four objectives: confidence, proximity, sparsity and plausibility (see Section~\ref{sec:related_work_cfs}). Following \citet{refoyo_sub-space_2024}, we add validity as a constraint rather than an objective. This results in the following optimization problem:
\begin{equation}
\begin{aligned}
\label{obj:optim}
    \min_{S_{\mathbf{x}_{\text{CF}}}} \text{ } & \alpha_\text{conf}\cdot O_\text{conf} +\alpha_\text{prox}\cdot O_\text{prox} + \alpha_\text{s}\cdot O_\text{s} +\alpha_\text{pl}\cdot O_\text{pl} \\
    \text{s.t. } & y_{\mathbf{x}_\text{CF}}\neq y_{\mathbf{x}}, 
\end{aligned}
\end{equation}
where the objectives are combined into a single objective using weights $\alpha_i$. In the remainder of this section, we define the objectives in the optimization problem. 


For the confidence objective $O_\text{conf}$, we maximize the output probability of the highest predicted class, excluding the originally predicted class $y_\mathbf{x}$. Mathematically, we denote $p_i(\mathbf{x}_{\text{CF}})$ as the output probability for class $i$, and define $O_\text{conf}$ as:
\begin{equation*}
    O_\text{conf}(\mathbf{x}, \mathbf{x}_{\text{CF}}) = 1 - \max_{\substack{i\in\{1, ..., n_c\} \\ i\neq y_\mathbf{x}}}p_i(\mathbf{x}_{\text{CF}}).
\end{equation*}

For the proximity objective $O_\text{prox}$, we measure the point-based proximity between the original sample and the counterfactual using the L1-norm, based on~\citet{wachter_counterfactual_2017}:
\begin{equation*}
    O_\text{prox}(\mathbf{x}, \mathbf{x}_{\text{CF}}) = \frac{1}{n}\sum_{i=1}^n \frac{|x_i-x_{\text{CF}_i}|}{\text{MAD}_i},
\end{equation*}
where $\text{MAD}_i$ is the mean absolute deviation of time stamp $i$ in the training data.

For the sparsity objective $O_\text{s}$, we determine the fraction of concepts $c_i\in S_\mathbf{x}$ that are transformed to obtain $S_{\mathbf{x}_{\text{CF}}}$, that is:
\begin{equation*}
    O_\text{s}(S_{\mathbf{x}}, S_{\mathbf{x}_{\text{CF}}}) = \frac{1}{m}\sum_{i=1}^m \mathds{1}_{c_i\neq c_{\text{CF}_i}}.
\end{equation*}

Finally, the plausibility objective $O_\text{pl}$ is defined based on the reconstruction error of a trained autoencoder $f_\text{AE}(\cdot)$. However, rather than directly minimizing the reconstruction error of the counterfactual, we follow \citet{refoyo_sub-space_2024} and define the objective as the increment in outlier score (IOS). Minimizing the reconstruction error directly is not desirable if the original sample already has a high reconstruction error, as this might drive the counterfactual into a low-error region that is semantically different from the original. Given the maximum reconstruction score for the training dataset $e_\text{max}$, $O_\text{pl}$ is defined as:
\begin{equation}
\label{eq:plausibility}
    O_\text{pl}(\mathbf{x}, \mathbf{x}_\text{CF}) = \frac{e(\mathbf{x}, \mathbf{x}_\text{CF})}{e_\text{max}},
\end{equation}
where:
\begin{equation*}
    e(\mathbf{x}, \mathbf{x}_\text{CF}) = \min(0, ||\mathbf{x}_\text{CF} - f_\text{AE}(\mathbf{x}_\text{CF})||_2 - ||\mathbf{x} - f_\text{AE}(\mathbf{x})||_2).
\end{equation*}

\subsubsection{Targeted counterfactuals}
Alternatively, rather than flipping the class, the optimization problem can be targeted towards generating a counterfactual of a specific class $y_\text{target} \neq y_\mathbf{x}$. In this case the validity constraint becomes: $y_{\mathbf{x}_\text{CF}} =  y_\text{target}$ and the confidence objective becomes:
\begin{equation*}
    O_\text{conf}(\mathbf{x}, \mathbf{x}_{\text{CF}}) = 1 - p_{y_{\text{target}}}(\mathbf{x}_{\text{CF}}).
\end{equation*}

\section{Implementation}
\label{sec:implementation}

In this section, we propose an approach to concept construction for ConceptCF, and discuss how to solve the concept-based counterfactual problem using a genetic algorithm. For the concept construction, we use three time series decomposition algorithms since these decompositions are invertible.

\subsection{Concept construction}

\paragraph{Discrete Fourier Transform} 
The first approach is the frequency-based approach used by TSEvo~\cite{hollig_tsevo_2022}, where a signal is decomposed into frequency bands using the Discrete Fourier Transform (DFT). The DFT decomposes a signal $\mathbf{x}$ as sinusoids parametrized by complex coefficients $\{a_0, a_1, ..., a_{n-1}| a_j\in\mathds{C}\}$, where each $x_k\in \mathbf{x}$ is represented by:
\begin{equation*}
     x_k= \sum_{j=0}^{n-1} a_j e^{-2i\pi \frac{kj}{n}}.
\end{equation*}
Following~\citet{hollig_tsevo_2022}, we define sets of frequency bands quadratically increasing in size. Frequency bands are given by $A_j = \{a_{j_s}, \dots, a_{j_e}\}$, with $j_s=\sum_{k=0}^{j-1} k^2$ and $j_e = \min\left(j_s + j^2, n\right)$. The last frequency band $A_B$ contains the highest remaining frequencies. Accordingly, we define our set of concepts as $S_\mathbf{x} = \{A_1, ..., A_B\}$. 

\paragraph{Discrete Wavelet Transform} 
The DFT may result in non-physically interpretable concepts if data is not strictly periodic. The Discrete Wavelet Transform (DWT)~\cite{heil1989continuous} was proposed as an alternative. The DWT filters a signal into an approximation component ($\mathbf{A}\in\mathds{R}^n$) and $n_\text{detail}$ detail components ($\mathbf{D}_i\in\mathds{R}^n$) using localized wavelet functions, i.e.:
\begin{equation*}
    \mathbf{x} = \mathbf{A} + \sum_{i=1}^{n_\text{detail}} \mathbf{D}_i.
\end{equation*}
Hence, using the DWT, we can define the set of concepts $S_\mathbf{x} = \{\mathbf{A}, \mathbf{D}_1, ..., \mathbf{D}_{n_\text{detail}}\}$.

\paragraph{Human-centered Decomposition} 
The DWT still does not ensure interpretable concepts, especially for end users who may not be domain experts. Hence, concepts are also constructed using the human-centered decomposition proposed by~\citet{jutte_c-shap_2025}. This decomposition decomposes a sample in a number of components with semantic meaning. Firstly, the decomposition extracts: $c_\text{Trend}$ as the linear trend, $c_\text{Bias}$ as the mean, and $c_\text{Scale}$ as the maximum amplitude of the sample. Given time stamps $\mathbf{t}\in \mathds{R}^n$, these can be represented as scalars in $\mathds{R}$. Additionally, three sequential components are extracted: $\mathbf{c}_\text{LF}$ as low frequency components extracted using a wavelet transform, $\mathbf{c}_\text{Var}$ as the sliding moving variance, and $\mathbf{c}_\text{HF}$ as the remaining high frequency component after filtering. This results in the following decomposition:
\begin{equation*}
    \mathbf{x} = c_\text{Trend} \cdot \mathbf{t} + c_\text{Bias} + c_\text{Scale} \cdot \left(\mathbf{c}_\text{LF} + \mathbf{c}_\text{Var} \cdot \mathbf{c}_\text{HF}\right).
\end{equation*}

For more details we refer to~\citet{jutte_c-shap_2025}. This decomposition results in the set of concepts $S_\mathbf{x} = \{c_\text{Trend}, c_\text{Bias}, c_\text{Scale}, \mathbf{c}_\text{LF}, \mathbf{c}_\text{Var}, \mathbf{c}_\text{HF}\}$.

\subsection{Optimization}
For the optimization of Eq.~\ref{obj:optim}, we use genetic evolution since it has shown to be effective in the state-of-the-art~\cite{dandl_multi-objective_2020} and allows for the non-differentiable sparsity objective.  For the genetic optimization, we represent each individual by a set of concepts $S_{\mathbf{x}_\textbf{CF}}$. Following~\citet{refoyo_sub-space_2024}, we turn the optimization problem into a fitness function and add the validity constraint to the function with penalty parameter $\beta$:
\begin{equation}
\begin{aligned}
    F(S_{\mathbf{x}_{\text{CF}}}) = &-\alpha_\text{conf} \cdot O_\text{conf} -\alpha_\text{prox} \cdot O_\text{prox} \\ &- \alpha_\text{s} \cdot O_\text{s} -\alpha_\text{pl}  \cdot O_\text{pl} - \beta \cdot \mathds{1}_{y_{\mathbf{x}_\text{CF}} = y_{\mathbf{x}}},
\end{aligned}
\label{eq:fitness}
\end{equation}
where the genetic algorithm should maximize the fitness function. Note, in case of a targeted counterfactual, the validity term becomes $\beta \cdot \mathds{1}_{y_{\mathbf{x}_\text{CF}} \neq y_{\text{target}}}$.

\subsubsection{Initial population}
To construct the initial population, we use the nearest unlike neighbour (NUN)~\cite{nugent_gaining_2009} of the input sample. The NUN $\mathbf{x}_\text{NUN}$ is defined as the most similar sample to $\mathbf{x}$ in the training data for which the class label differs ($y_\mathbf{x} \neq y_{\mathbf{x}_\text{NUN}}$). Similarity is measured in the sample space (point-based) using the Euclidean norm. Each individual in the initial population is initialized to $\mathbf{x}_\text{NUN}$, after which mutation (as will be described shortly) is applied to all individuals except one. Consequently, the NUN is assured to be part of the initial population.

For targeted counterfactuals, the NUN is considered to be the nearest neighbour for which $y_{\mathbf{x}_\text{NUN}} = y_\text{target}$.

\subsubsection{Parent selection and crossover}
For mating, $n_\text{parents}$ are selected through tournament selection, due to its established efficiency~\cite{shukla_comparative_2015}. The offspring is obtained by single-point crossover of the concepts $c_i \in S_{\mathbf{x}_\text{CF}}$. Additionally, the top $n_\text{elite} \geq 1$ individuals from the current generation are carried to the next generation.

\subsubsection{Mutation}
Mutation is applied independently to each concept $c_i \in S_{\mathbf{x}_\text{CF}}$ with probability $p_\text{mutation}$. We can have both scalar and sequential concepts which require different mutation approaches.

For scalar concepts, the concept value $c_i$ is perturbed by random noise. Given variance $\sigma_{c_i}$ of the concept in the training data, noise is sampled from $N(0, \sigma_{c_i})$. Resulting values exceeding the observed range of the concept in the training data $(\left[c^{\text{min}}_i, c^{\text{max}}_i\right]$) are clipped to the nearest boundary.

Sequential concepts require a mutation approach that preserves semantic meaning, this would not be ensured when applying point-based random noise. For example, a concept such as `Low frequency' models repetitive slow changes over time. When noise is applied to individual points within the concept, this may introduce quick changes, contradicting its interpretation. Consequently, rather than perturbing individual points, we replace the full sequence with an alternative drawn from the training data, similar to the CoMTE~\cite{ates_counterfactual_2021} approach for multivariate data. 

Specifically, for each sequential concept, we build a KDtree of all sequences in the training set (using the Euclidean distance). From the tree, we retrieve the $n_\text{NN}$ nearest neighbours to the current concept. From these, one sample is selected according to probabilities inversely proportional to the distance to the current concept. This strategy prioritises exploration of similar solutions (for the proximity objective) before expanding to more distant solutions.

Finally, for the mutation, to facilitate sparsity, with probability $p_\text{reset}$ concepts are set to the original concepts in $S_\mathbf{x}$, as suggested by \citet{dandl_multi-objective_2020}. 

\subsubsection{Validity and optimization bound}
We conclude this section by establishing two properties of ConceptCF. We establish them for the non-targeted variant. Their extension to the targeted variant is trivial. 

By definition, the NUN $\mathbf{x}_\text{NUN}$ belongs to a different class from $\mathbf{x}$. Hence, it satisfies the validity constraint. When the penalty weight $\beta$ is sufficiently large, any samples not matching the validity constraint will result in a lower fitness. Since the NUN is included in the initial population and the fittest individuals are preserved across generations via elitism, no generation can produce a best solution with lower fitness than the NUN. Thus, the best individual in every generation satisfies the validity constraint and has higher fitness than the NUN. We pose this as the following proposition:
\begin{proposition}
For a counterfactual $\mathbf{x}_\text{CF}$ generated using ConceptCF, given nearest unlike neighbour $\mathbf{x}_\text{NUN}$ and $\beta$ sufficiently large, we have:
\begin{enumerate}
    \item $y_{\mathbf{x}} \neq y_{\mathbf{x}_\text{CF}}$, i.e. valid solutions for Eq.~\ref{obj:optim} are guaranteed.
    \item $F(S_{\mathbf{x}_{\text{CF}}}) \geq F(S_{\mathbf{x}_{\text{NUN}}})$, i.e. $S_{\mathbf{x}_\text{NUN}}$ is the worst case solution.
\end{enumerate}
\end{proposition}

\section{Experimental setup}
\label{sec:experiments}

In this section, we describe the implementation of ConceptCF and the experiments conducted. For the implementation Python 3.9 and PyTorch 2.8 were used to match the packages for the baseline methods. The experiments were run on an AMD Ryzen Threadripper 2950X 16-Core Processor (62 GB RAM) in Ubuntu 22.04.5 LTS with an NVIDIA TITAN V GPU (CUDA Version: 12.8).

\subsection{Data}
The experiments are conducted on the MotionSenseHAR (MS-HAR) dataset from the UEA archive~\cite{bagnall_uea_2018} and four datasets from the UCR archive~\cite{dau_ucr_2019}: CBF, Coffee, ECG200, GunPoint. All datasets are normalized using the mean and standard deviation computed across all training samples and temporal points.

MS-HAR is a multivariate dataset. We convert it to univariate by using only the accelerometer measurements along the z-axis. Additionally, the original sequences of length 1,000 are split into four samples of length 250 (4.98 s) to reduce computation time. No further processing is applied to the other four datasets.

\subsection{Predictive models}

For the black box models, we follow~\citet{delaney_instance-based_2021}, and use the fully convolutional neural network (FCN) architecture by~\citet{wang_time_2017}. The models consist of three 1D convolutional layers, combined with batch normalization, followed by a global average pooling layer, and a fully connected layer. 

\subsection{Concept hyperparameters}
For the concept construction using the DFT, no hyperparameters are used. For the DWT, the Daubechies wavelet with six vanishing moments is used to extract four components (i.e. $n_{\text{detail}}=3$). For the human-centered decomposition, the implementation and hyperparameters in \citet{jutte_c-shap_2025} are used, where the sliding window size for the extraction of the variance component is set to 27.

\subsection{Optimization}
For the optimization, hyperparameters are chosen based on initial experimentation. The weights of the fitness function (see Eq.~\ref{eq:fitness}) are set to: $\alpha_\text{conf}=0.4$, $\alpha_\text{prox}=0.6$, $\alpha_\text{s}=0.4$, $\alpha_\text{pl}=0.2$ and $\beta = 100$. 

For the genetic algorithm, the PyGAD library~\cite{gad2023pygad} is used. The algorithm is run for $n_\text{gen} = 100$ generations with a population size of $n_\text{indiv} = 50$. The best $n_\text{elite} = 2$ solutions in a generation are copied to the next generation. If no improvement in fitness is found for $n_\text{saturate}=15$ generations, the algorithm is terminated. For the mating, $n_\text{parents} = 10$ are used. Concepts are mutated with $p_\text{mutation} = 0.2$ and $p_\text{reset} = 0.05$. For the sampling of sequential concepts, $n_\text{NN}=50$ neighbours are considered.

For the plausibility objective, we replicate the autoencoder architecture from Glacier~\cite{wang_glacier_2024}. To reduce computational costs, we limit the test and reference training data for the counterfactual generation to 100 random samples per dataset.

\subsection{Baselines}
As baselines, we consider five methods from literature: Wachter~\cite{wachter_counterfactual_2017}, Native Guide~\cite{delaney_instance-based_2021}, TSEvo~\cite{hollig_tsevo_2022}, Glacier~\cite{wang_glacier_2024}, and Sub-SpaCE~\cite{refoyo_sub-space_2024}. For descriptions of these methods, see Section~\ref{sec:related_work_cfs}.

For implementation of TSEvo, the TSInterpret~\cite{hollig2023tsinterpret} library is used. For the other baselines, the Counterfactual Explanation Algorithms for Time Series Models (CFTS)~\cite{schlegel_what-if_2026} library is used. The Glacier and Sub-SpaCE implementations are adapted to include autoencoders, using the same as trained for ConceptCF. Additionally, for fair comparison, early stopping is disabled in CFTS. TSEvo is run for 100 generations. All other settings use the defaults in the packages.

\subsection{Metrics}
\label{sec:metrics}
For the metrics we follow~\citet{refoyo_sub-space_2024} and consider metrics in line with the previously defined objectives for the optimization problem:
\begin{itemize}
    \item Validity: The percentage of counterfactuals that change the original output class.
    \item Confidence: The confidence of the predictive model in its prediction of the counterfactual.
    \item Proximity: The pointwise root mean squared error between the original sample and the counterfactual.
    \item Sparsity: The fraction of `meaningful' units changed from the original sample to the counterfactual. For Native Guide and Sub-SpaCE we consider the number of subsequences changed. For ConceptCF, we consider the number of concepts. For the other approaches, we consider the number of points. 
    \item Plausibility: The increase in outlier score (IOS)~\cite{refoyo_sub-space_2024}, see Eq.~\ref{eq:plausibility}.
\end{itemize}

\section{Results}
\label{sec:results}

\subsection{Demonstration}
\begin{figure*}[!htbp]
   \begin{subfigure}{0.33\textwidth}
        \centering
        \includegraphics[width=\linewidth]{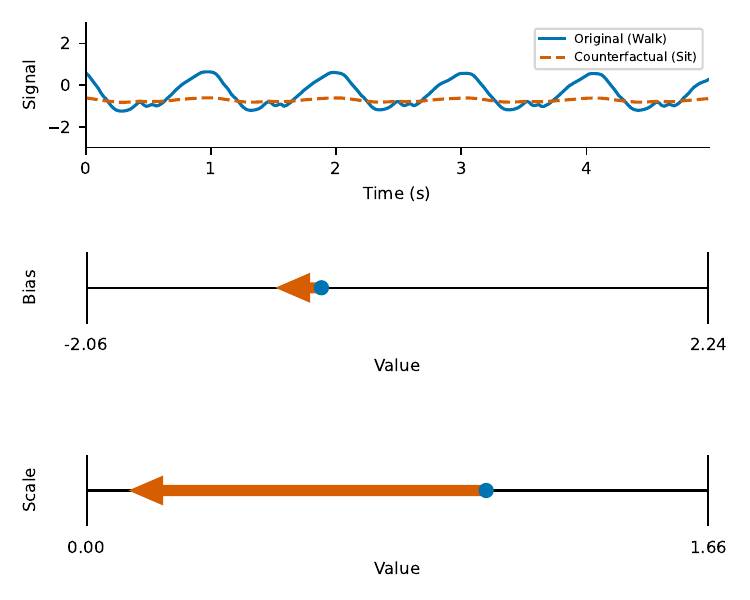}
        \caption{ConceptCF: Human-centered}
        \label{fig:ex_cust_dec}
  \end{subfigure}
  \hfill
  \begin{subfigure}{0.33\textwidth}
        \centering
        \includegraphics[width=\linewidth]{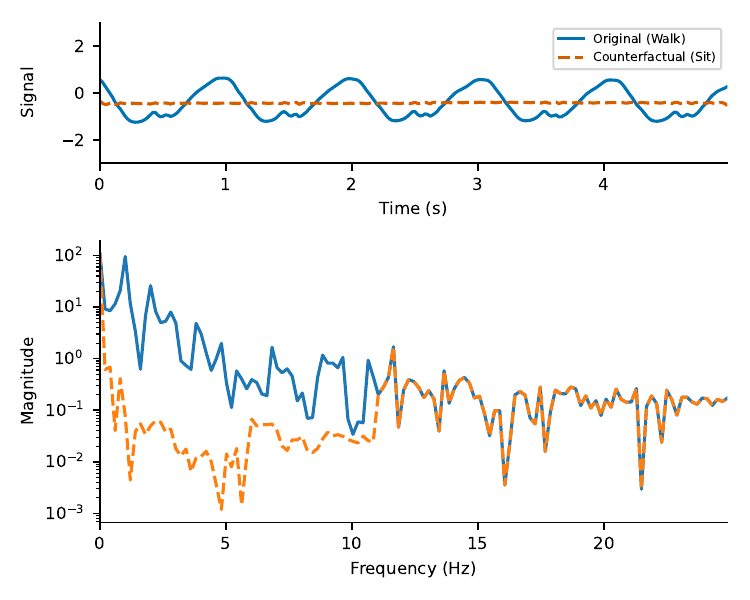}
        \caption{ConceptCF: DFT}
        \label{fig:fft_ex}
  \end{subfigure}
  \hfill
  \begin{subfigure}{0.33\textwidth}
      \centering
      \includegraphics[width=\linewidth]{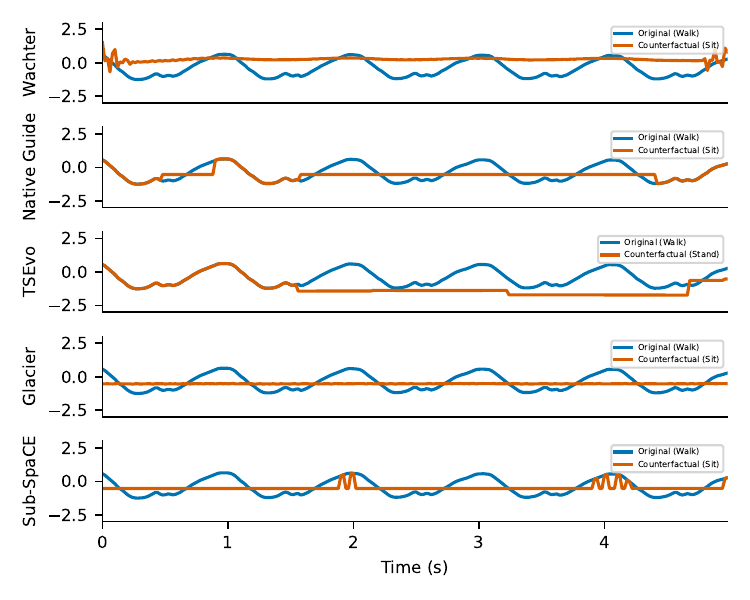}
      \caption{Baselines}
      \label{fig:ex_baselines}
  \end{subfigure}
\caption{Counterfactual explanations for a selected sample from the MS-HAR dataset of class `Walk', counterfactuals are generated for class `Sit'. Note, for the human-centered decomposition only the changed concepts are visualized.}
\label{fig:example}
\end{figure*}

As demonstration, we apply ConceptCF and the baselines to a sample of class `Walk' from the MS-HAR dataset, shown in Fig.~\ref{fig:example}. Targeted generation was applied towards class `Sit'. The human-centered decomposition shows that the model would classify the sample as `Sit' rather than `Walk' if the `Scale' (i.e. amplitude) of the signal was lower and the `Bias' (i.e. orientation) was shifted. The DFT approach shows that the low frequency components differ between the two classes. For the baselines, although all methods suggest minimizing fluctuations in (parts of) the signal, it cannot directly be seen whether all frequency components were modified. Additionally, any changes in `Bias' are less apparent.

\subsection{Evaluation}

For the evaluation, counterfactuals are generated in a non-targeted manner. Inspired by~\citet{cetina_counterfactual_2026}, we rank the approaches. First, we rank per dataset using pairwise one-tailed Wilcoxon signed-rank tests ($\alpha = 0.05$) on the sample-level scores. An approach receives a lower rank if it is significantly worse than any in the rank above. Second, we rank globally by applying the same procedure to the mean scores per dataset. This latter ranking should be interpreted cautiously due to the small sample (5 datasets).

Table~\ref{tab:decomposition} shows the results for the different decomposition approaches. For readability, only the highest rank is indicated on the dataset-level. Validity is not included since ConceptCF always results in valid solutions (cf. Proposition 1). The three approaches score comparably across metrics, with only the DWT-based performing worse with respect to sparsity.

\begin{table}[htb]

\caption{Results of ConceptCF for the three approaches: Human-centered (H-C), Discrete Fourier Transform (DFT), and Discrete Wavelet Transform (DWT). Scores are reported as means over the evaluation data. Methods are ranked based on pairwise statistical significance ($\alpha = 0.05$). The best scores per dataset are put in bold font.}

\vspace{2mm}

\centering
\begin{small}
\begin{tabular}{@{}lccccc@{}}
\multicolumn{6}{c}{Confidence ($\uparrow$)} \\ \toprule

  & MS-HAR & CBF & Coffee & ECG200 & GunPoint \\ \midrule
\textbf{H-C (1)} & 0.933 & \textbf{0.730} & \textbf{0.643} & \textbf{0.857} & 0.819 \\
\textbf{DFT (1)} & 0.907 & 0.643 & \textbf{0.661} & 0.823 & \textbf{0.874} \\
\textbf{DWT (1)} & \textbf{0.945} & \textbf{0.751} & 0.621 & \textbf{0.844} & \textbf{0.874} \\ \bottomrule \\
\end{tabular}
\end{small}

\begin{small}
\begin{tabular}{@{}lccccc@{}}
\multicolumn{6}{c}{Proximity ($\downarrow$)} \\ \toprule
 & MS-HAR & CBF & Coffee & ECG200 & GunPoint \\ \midrule
\textbf{H-C (1)} & \textbf{0.189} & \textbf{0.540} & 0.060 & 0.316 & \textbf{0.246} \\
\textbf{DFT (1)} & \textbf{0.321} & \textbf{0.526} & 0.046 & \textbf{0.287} & \textbf{0.208} \\
\textbf{DWT (1)} & 0.527 & 0.663 & \textbf{0.043} & \textbf{0.295} & \textbf{0.252} \\ \bottomrule \\
\end{tabular}
\end{small}

\begin{small}
\begin{tabular}{@{}lccccc@{}}
\multicolumn{6}{c}{Sparsity ($\downarrow$)} \\ \toprule

  & MS-HAR & CBF & Coffee & ECG200 & GunPoint \\ \midrule
\textbf{H-C (1)} & 0.288 & 0.283 & 0.179 & \textbf{0.200} & 0.330 \\
\textbf{DFT (1)} & \textbf{0.199} & \textbf{0.235} & \textbf{0.134} & 0.266 & \textbf{0.282} \\
DWT (2) & 0.323 & 0.255 & 0.268 & 0.338 & 0.335 \\
\bottomrule \\
\end{tabular}
\end{small}

\begin{small}
\begin{tabular}{@{}lccccc@{}}
\multicolumn{6}{c}{Plausibility ($\downarrow$)} \\ \toprule
 & MS-HAR & CBF & Coffee & ECG200 & GunPoint \\ \midrule
\textbf{H-C (1)} & \textbf{0.721} & \textbf{0.753} & \textbf{0.797} & \textbf{0.623} & \textbf{0.417} \\
\textbf{DFT (1)} & \textbf{0.575} & 0.845 & 0.992 & 0.759 & 0.483 \\
\textbf{DWT (1)} & \textbf{0.471} & 0.843 & 1.005 & 0.759 & 0.492 \\ \bottomrule
\end{tabular}
\end{small}

\label{tab:decomposition}
\end{table}

For the baselines, the methods reached 100\% validity on all datasets, except Wachter which reached 92.9\% validity on the Coffee dataset. The other results can be found in Table~\ref{tab:confidence}. For ConceptCF, only the results from the human-centered approach are included. 

\begin{table}[!htb]

\caption{Results for ConceptCF (C-CF), Wachter (W), Native Guide (NG), TSEvo, Glacier (Gla) and Sub-SpaCE (Sub-S). Scores are reported as means over the evaluation data. Methods are ranked based on pairwise statistical significance ($\alpha = 0.05$). The best scores per dataset are put in bold font.
}

\vspace{2mm}

\centering

\begin{small}
\begin{tabular}{@{}lcccccc@{}}
\multicolumn{7}{c}{Confidence ($\uparrow$)} \\ \toprule
 & \begin{tabular}[c]{@{}c@{}}C-CF\\ (2)\end{tabular} & \begin{tabular}[c]{@{}c@{}}W\\ (2)\end{tabular} & \begin{tabular}[c]{@{}c@{}}NG\\ (3)\end{tabular} & \begin{tabular}[c]{@{}c@{}}TSEvo\\ (4)\end{tabular} & \begin{tabular}[c]{@{}c@{}}\textbf{Gla}\\ \textbf{(1)}\end{tabular} & \begin{tabular}[c]{@{}c@{}}\textbf{Sub-S}\\ \textbf{(1)}\end{tabular} \\ \midrule
MS-HAR & 0.933 & 0.933 & 0.570 & 0.259 & \textbf{0.992} & 0.822 \\
CBF & 0.730 & 0.833 & 0.514 & 0.336 & 0.925 & \textbf{0.944} \\
Coffee & 0.643 & 0.608 & 0.586 & 0.500 & \textbf{0.970} & 0.942 \\
ECG200 & 0.857 & 0.899 & 0.664 & 0.500 & 0.986 & \textbf{0.998} \\
GunPoint & 0.819 & 0.880 & 0.649 & 0.500 & \textbf{0.987} & 0.944 \\ \bottomrule \\
\end{tabular}

\begin{tabular}{@{}lcccccc@{}}
\multicolumn{7}{c}{Proximity ($\downarrow$)} \\ \toprule
& \begin{tabular}[c]{@{}c@{}}\textbf{C-CF}\\ \textbf{(1)}\end{tabular} & \begin{tabular}[c]{@{}c@{}}\textbf{W}\\ \textbf{(1)}\end{tabular} & \begin{tabular}[c]{@{}c@{}}\textbf{NG}\\ \textbf{(1)}\end{tabular} & \begin{tabular}[c]{@{}c@{}}TSEvo\\ (2)\end{tabular} & \begin{tabular}[c]{@{}c@{}}Gla\\ (2)\end{tabular} & \begin{tabular}[c]{@{}c@{}}\textbf{Sub-S}\\ \textbf{(1)}\end{tabular} \\ \midrule
MS-HAR & \textbf{0.189} & 0.766 & 0.540 & 0.987 & 0.934 & 0.665 \\
CBF & 0.540 & \textbf{0.231} & 0.567 & 0.645 & 0.849 & 0.785 \\
Coffee & 0.060 & \textbf{0.018} & 0.074 & 0.075 & 0.118 & 0.081 \\
ECG200 & 0.316 & \textbf{0.068} & 0.297 & 0.374 & 0.516 & 0.365 \\
GunPoint & 0.246 & \textbf{0.044} & 0.222 & 0.243 & 0.309 & 0.214 \\ \bottomrule \\
\end{tabular}

\begin{tabular}{@{}lcccccc@{}}
\multicolumn{7}{c}{Sparsity ($\downarrow$)} \\ \toprule
& \begin{tabular}[c]{@{}c@{}}\textbf{C-CF}\\ \textbf{(1)}\end{tabular} & \begin{tabular}[c]{@{}c@{}}W\\ (4)\end{tabular} & \begin{tabular}[c]{@{}c@{}}NG\\ (2)\end{tabular} & \begin{tabular}[c]{@{}c@{}}\textbf{TSEvo}\\ \textbf{(1)}\end{tabular} & \begin{tabular}[c]{@{}c@{}}Gla\\ (4)\end{tabular} & \begin{tabular}[c]{@{}c@{}}Sub-S\\ (3)\end{tabular} \\ \midrule
MS-HAR & \textbf{0.288} & 1.000 & 0.366 & 0.589 & 1.000 & 0.405 \\
CBF & \textbf{0.283} & 1.000 & 0.352 & 0.353 & 1.000 & 0.445 \\
Coffee & \textbf{0.262} & 1.000 & 0.364 & 0.237 & 1.000 & 0.465 \\
ECG200 & \textbf{0.200} & 1.000 & 0.355 & 0.264 & 1.000 & 0.475 \\
GunPoint & \textbf{0.330} & 1.000 & 0.348 & \textbf{0.312} & 1.000 & 0.449 \\ \bottomrule \\
\end{tabular}

\begin{tabular}{@{}lcccccc@{}}
\multicolumn{7}{c}{Plausibility ($\downarrow$)} \\ \toprule
& \begin{tabular}[c]{@{}c@{}}C-CF\\ (2)\end{tabular} & \begin{tabular}[c]{@{}c@{}}W\\ (3)\end{tabular} & \begin{tabular}[c]{@{}c@{}}NG\\ (2)\end{tabular} & \begin{tabular}[c]{@{}c@{}}TSEvo\\ (3)\end{tabular} & \begin{tabular}[c]{@{}c@{}}\textbf{Gla}\\ \textbf{(1)}\end{tabular} & \begin{tabular}[c]{@{}c@{}}Sub-S\\ (3)\end{tabular} \\ \midrule
MS-HAR & 0.721 & 2.636 & 0.773 & 1.271 & \textbf{0.235} & 1.353 \\
CBF & 0.753 & 0.854 & 0.732 & 0.836 & \textbf{0.478} & 0.682 \\
Coffee & \textbf{0.797} & 0.973 & 0.943 & 1.101 & \textbf{0.777} & 0.953 \\
ECG200 & 0.623 & 0.913 & 0.738 & 0.879 & \textbf{0.524} & 1.295 \\
GunPoint & 0.417 & 0.673 & 0.580 & 0.857 &\textbf{0.317} & 1.076 \\ \bottomrule
\end{tabular}

\end{small}

\label{tab:confidence}

\end{table}

ConceptCF achieves the highest rank for the proximity and sparsity metrics. While Wachter outperforms ConceptCF on four of the five datasets, it cannot optimize for sparsity. ConceptCF has the best sparsity across all datasets, with only TSEvo achieving comparable results on the GunPoint dataset. On the confidence and plausibility metrics, ConceptCF ranks second. Only Glacier performs better, it has an edge in plausibility by directly generating counterfactuals in the autoencoders' latent space.

In conclusion, ConceptCF demonstrates competitive performance compared to state-of-the-art methods, with particular strength in sparsity. By requiring few changes to construct a counterfactual, ConceptCF produces explanations that are easier for users to interpret, aligning with its primary objective.

\subsection{Sensitivity analysis}

\begin{figure}[!htb]
    \centering
    \begin{subfigure}{0.48\linewidth}
        \centering
        \includegraphics[width=\linewidth]{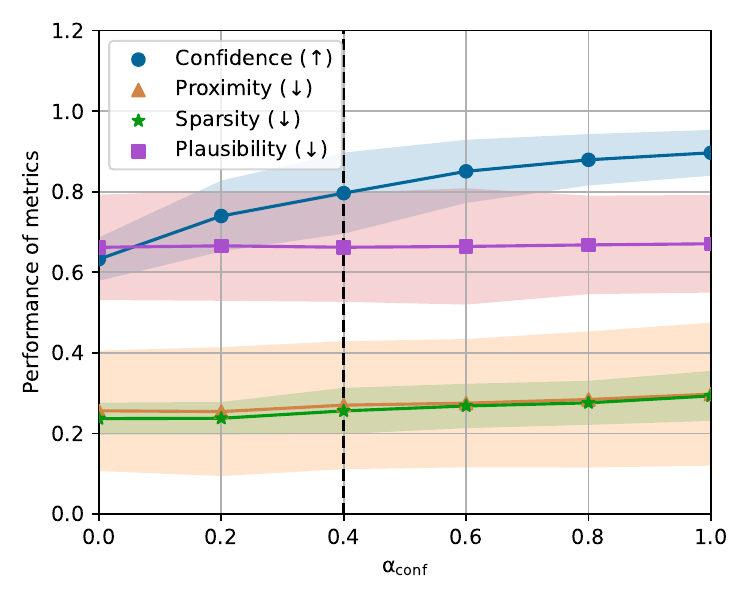}
        \caption{Confidence}
    \end{subfigure}
    \begin{subfigure}{0.48\linewidth}
        \centering
        \includegraphics[width=\linewidth]{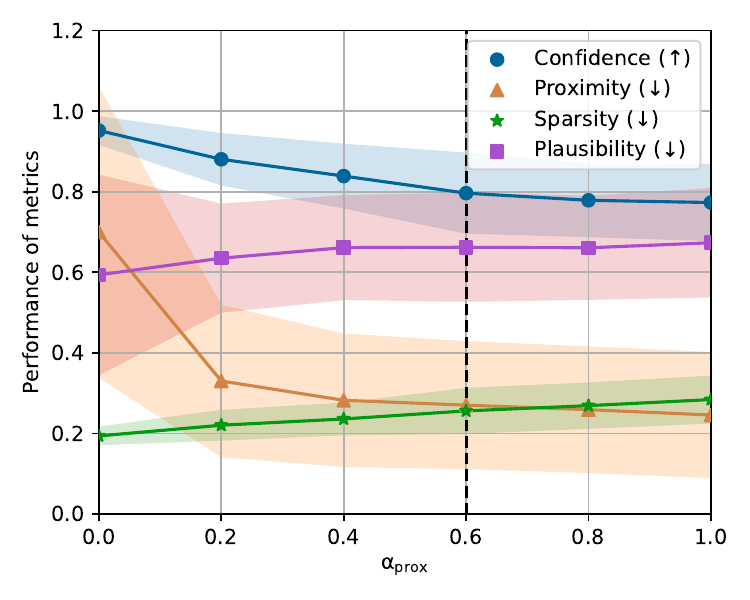}
        \caption{Proximity}
    \end{subfigure}

        \begin{subfigure}{0.48\linewidth}
        \centering
        \includegraphics[width=\linewidth]{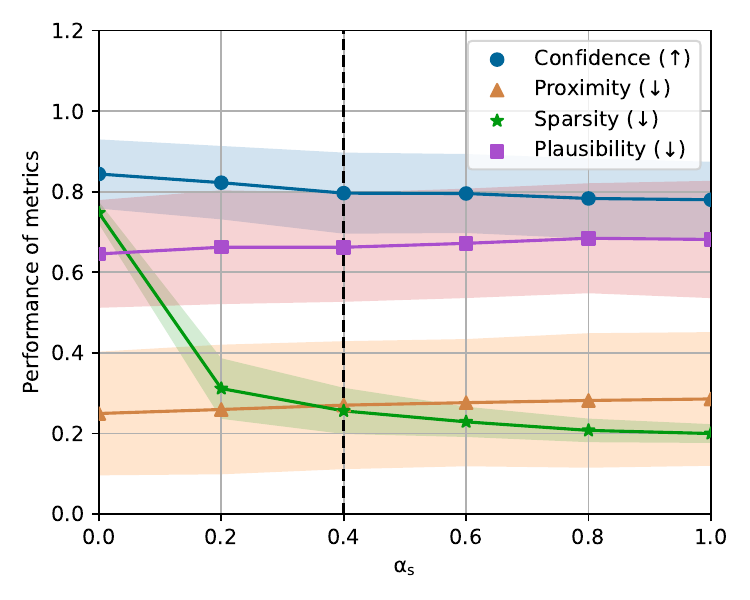}
        \caption{Sparsity}
    \end{subfigure}
    \begin{subfigure}{0.48\linewidth}
        \centering
        \includegraphics[width=\linewidth]{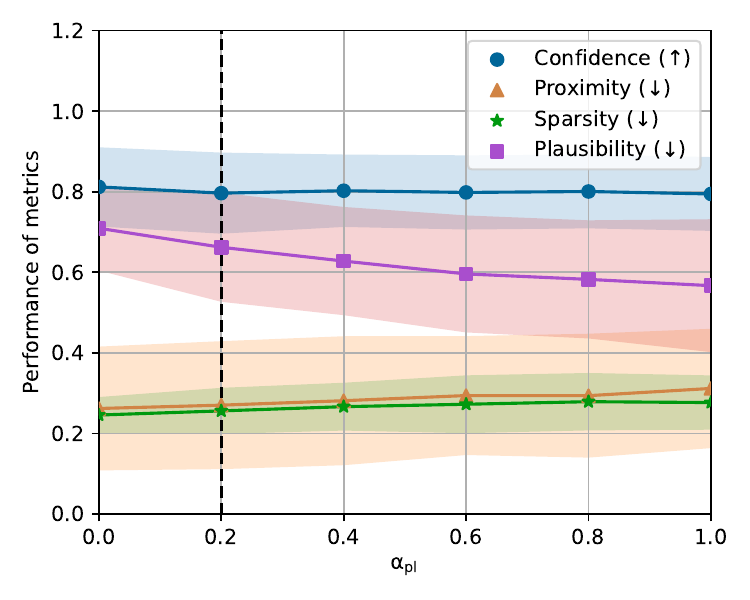}
        \caption{Plausibility}
    \end{subfigure}
    \caption{Sensitivity analysis of the objective weights $\alpha_i$ with respect to the above metrics. The mean and standard deviation over the five datasets are shown. The vertical lines indicate the original values of $\alpha_i$.}
    \label{fig:sensitivity}
\end{figure}

The sensitivity analysis evaluates the effects of the objective weights $\alpha_\text{conf}$, $\alpha_\text{prox}$, $\alpha_\text{s}$ and $\alpha_\text{pl}$ (Eq.~\ref{obj:optim}) on the four evaluation metrics: confidence, proximity, sparsity and plausibility. Each weight is varied independently across the values $[0.0, 0.2, 0.4, 0.6, 0.8, 1.0]$ while all other hyperparameters remain fixed. The results can be found in Fig.~\ref{fig:sensitivity}. Validity is omitted because it remains at 100\%.

Each objective's corresponding metric responds to weight adjustments, confirming tunability. Due to the balancing in the multi-objective problem, modifying one weight affects the other metrics. In practice, the choice should be adapted to the requirements of end users.

\section{Conclusions and future work}

In this paper, we introduced ConceptCF, an approach to counterfactual generation for time series. ConceptCF differs from existing methods by generating counterfactuals in terms of high-level human-interpretable concepts, instead of individual timestamps or subsequences. This allows for explanations of global patterns in human interpretable terms.

In this paper, we considered concepts constructed through time series decomposition. The choice of decomposition had limited effect on the qualitative metrics. In practice, the choice should be aligned with the end users. Depending on the domain, the choice between simpler (e.g. the human-centered decomposition) and more complex representations (e.g. the DFT) can be made. Furthermore, in future research, other types of concepts could be considered. For example, localized patterns, thereby unifying ConceptCF with subsequence-based approaches, or autocorrelation. Additionally, unsupervised concept construction could be explored. Important is that concepts can be modified in an interpretable manner. 

Our experimental results demonstrate competitive performance against five baseline methods with respect to validity, confidence, proximity, sparsity, and plausibility. These metrics do not fully capture user comprehension, therefore user evaluation is suggested for future research. However, we argue ConceptCF's design inherently addresses interpretability through human-understandable concepts. 

\section*{Acknowledgments}
This publication is part of the project ZORRO with project number KICH1.ST02.21.003 of the research programme Key Enabling Technologies (KIC) which is partly financed by the Dutch Research Council (NWO). This research is part of the SPRONG DEMAND. This research is partly financed by Taskforce for Applied Research SIA, part of the Dutch Research Council (NWO).

Generative AI, in the form of a local model hosted by the University of Twente, was used for code debugging, and to proofread and refine the grammar of this paper.

\newpage

\bibliographystyle{plainnat}
\bibliography{bibliography}

\end{document}